%% file: main.tex
\documentclass[runningheads]{llncs}
\usepackage{times}
\usepackage{epsfig}
\usepackage{graphicx}
\usepackage{amsmath,amssymb}
\usepackage{color}
\usepackage{latex_pkgs/maths}
\usepackage{latex_pkgs/msformatting}
\usepackage{algorithm2e}
\usepackage{listings}
\usepackage{cuted} 

\begin{document}

\title{Predicting Action Tubes} 
\titlerunning{Predicting Action Tubes}
%
\author{Gurkirt Singh \and Suman Saha \and Fabio Cuzzolin}
%
\authorrunning{G. Singh \etal}
%

\institute{Oxford Brookes University, UK\\
\email{gurkirt.singh-2015@brookes.ac.uk}}

\maketitle              
\begin{abstract}  
\input{text/abstract}
\end{abstract}
\input{text/intro}
\input{text/soa}
\input{text/approach}
\input{text/experiments_results_discussion}
\input{text/conclusion}
\clearpage
\bibliographystyle{splncs04}
\bibliography{bib}
\end{document}


\title{Supplementary Material: Predicting Action Tubes} 
\titlerunning{Predicting Action Tubes}
%
\author{Gurkirt Singh \and Suman Saha \and Fabio Cuzzolin}
%
\authorrunning{G. Singh \etal}
%

\institute{Oxford Brookes University, UK\\
\email{gurkirt.singh-2015@brookes.ac.uk}}

\maketitle              

\input{text/supp_details}
\clearpage
\bibliographystyle{splncs04}
\bibliography{bib}

%% file: text/abstract.tex
In this work, we present a method to predict an entire `action tube' (a set of temporally linked bounding boxes) in a trimmed video just by observing a smaller subset of it. 
Predicting where an action is going to take place in the near future is essential to many computer vision based applications such as autonomous driving or surgical robotics.
Importantly, it has to be done in real-time and in an online fashion.
We propose a \textbf{T}ube \textbf{P}rediction network (TPnet) which jointly predicts the past, present and future bounding boxes along with their action classification scores.  
At test time TPnet is used in a (temporal) sliding window setting, and its predictions are put into a tube estimation framework to construct/predict the video long action tubes not only for the observed part of the video but also for the unobserved part.
Additionally, the proposed action tube predictor helps in completing action tubes for unobserved segments of the video.
We quantitatively demonstrate 
the latter ability, and the fact that TPnet
improves state-of-the-art detection performance, on one of the standard action detection benchmarks - J-HMDB-21 dataset.

%% file: text/intro.tex
\section{Introduction} \label{sec:intro}

\begin{figure}[h]
  \centering
  \vspace{-0.6cm}
  \includegraphics[scale=0.63]{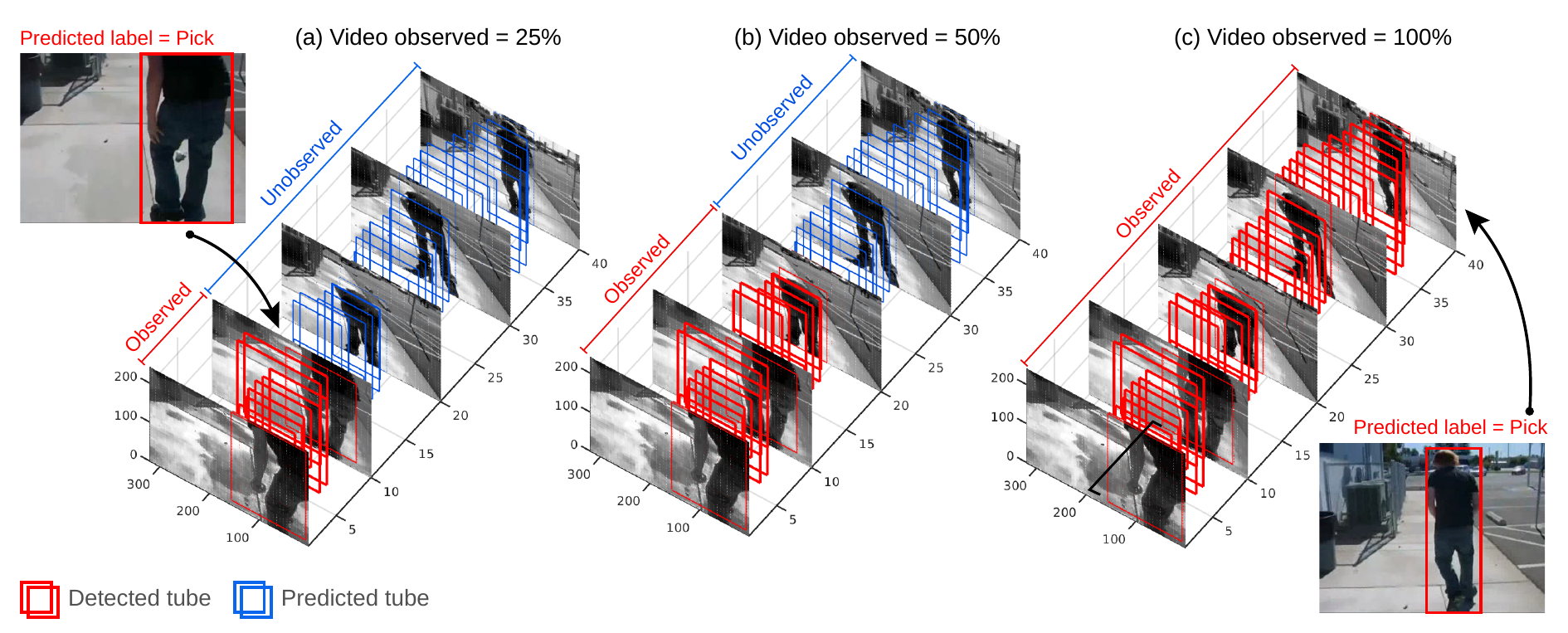}
  \caption{
    {\small
      \textit{
      An Illustration of the action tube prediction problem using an example in which a ``pickup'' action is being performed on a sidewalk.
      As an ideal case, we want the system to predict an action tube as shown in (c) (i.e. when 100\% of the video has been processed) just by observing 25\% of the entire clip (a).       
      We want the tube predictor to predict the action class label (shown in red) alongside predicting the spatial location of the tube.
        The red shaded bounding boxes denote the detected tube in the observed portion of the input video, whereas, the blue coloured bounding boxes represent the future predicted action tube for the unobserved part of the clip.        
      }
    }
    }
\label{fig:problem-statement}
\vspace{-6mm}
\end{figure}

Imagine a pedestrian on the sidewalk, and an autonomous car cruising on the nearby road. 
If the pedestrian stays on the sidewalk and continues walking, they are of no concern for the self-driving car.
What, instead, if they start approaching the road, in a possible attempt to cross it?
Any future prediction about the pedestrian's action and their possible position on/off the road would crucially help the autonomous car avoid any potential incident. It would suffice to foresee the pedestrian's action label and position half a second early to avoid a major accident.
As a result, awareness about surrounding human actions is essential for the robot-car.

We can formalise the problem as follows. We seek to predict both the class label and the future spatial location(s) of an action instance as early as possible, as shown in Figure~\ref{fig:problem-statement}. 
Basically, it translates into early spatiotemporal action detection~\cite{singh2016online}, achieved by completing action instance(s) for the unobserved part of the video.
As commonly accepted, action instances are here described by `tubes' formed by linking bounding box detections in time.

In an existing relevant work by Singh \etal~\cite{singh2016online}, 
early label prediction and online action detection are performed jointly. 
The action class label for an input video is predicted early on just 
by observing a smaller portion (a few frames) of it, whilst 
the system incrementally builds action tubes in an online fashion. 
In contrast, the proposed approach can predict both the class label of an action 
and its future location(s) (i.e., the future shape of an action tube). 
In this work, by \emph{`prediction'} we refer to the estimation of both 
an action's label and location in \emph{future}, unobserved video segments. 
We term \emph{`detection'} the estimation of action labels/locations in the 
observed segment of video up to any given time, i.e., 
for \emph{present} and \emph{past} video frames.

The computer vision community is witnessing a rising interest in problems such as 
early action label prediction~\cite{singh2016online,nazerfard2013using,hoai2014max,aliakbarian2017encouraging,Soomrocvpr2016,yeung2015every,yeungcvpr2016,kong2017deep,zunino2017predicting,ryoo2011human,lan2014hierarchical}, 
online temporal action detection ~\cite{ma2016learning,yeung2015every,de2016online,tahmidajoint},
online spatio-temporal action detection ~\cite{singh2016online,Soomrocvpr2016,yang2017spatio}, 
future representation prediction~\cite{vondrick2015anticipating,kong2017deep} or 
trajectory prediction ~\cite{alahi2016social,kitani2012activity,lee2017desire}. 
Although, all these problems are interesting, and definitely encompass a broad scope of applications, they do not entirely capture the complexity involved by many critical scenarios including, e.g., surgical robotics or autonomous driving.  
In opposition to \cite{singh2016online,Soomrocvpr2016}, which can only perform early label prediction and online action detection, in this work we propose to predict both future action location and action label.
A number of challenges make this problem particularly hard, e.g., the temporal structure of an action is obviously not completely observed; locating human actions is itself a difficult task; the observed part can only provide clues about the future locations. In addition,
camera movement can make it even harder to extrapolate an entire tube. 
We propose to solve these problems by regressing the future locations from the present tube.

The ability to predict \emph{action micro-tubes} (sets of temporally connected bounding boxes spanning $k$ video frames) from pairs of frames~\cite{saha2017amtnet} or sets of $k$ frames~\cite{kalogeiton2017action,hou2017tube} provides a powerful tool to extend the single frame-based online approach by Singh \etal~\cite{singh2016online} in order to cope with action location prediction, while retaining its incremental nature.
Combining the basic philosophies of \cite{singh2016online} and 
\cite{saha2017amtnet} has thus the potential to provide an interesting and scalable approach to action prediction. 

Briefly, the action micro-tubes network (AMTnet,~\cite{saha2017amtnet}),
divides the action tube detection problem into a set of smaller sub-problems. 
Action `micro-tubes' are produced by a convolutional neural network (a 3D region proposal network) processing two input frames that are $\Delta$ apart. Each micro-tube consists of two bounding boxes belonging to the two frames. When the network is applied to consecutive 
pairs of frames, it produces a set of consecutive micro-tubes which can be finally linked~\cite{singh2016online} to form complete action tubes.
The detections forming a micro-tube can be considered as implicitly linked, hence reducing the number of linking subproblems.
Whereas AMTnet was originally designed to generate micro-tubes using only appearance (RGB) inputs, here we augment it by
introducing the feature-level fusion of flow and appearance cues, drastically improving its performance and, as a result, that of TPnet.

\begin{figure*}[t]
  \centering
  \includegraphics[scale=0.60]{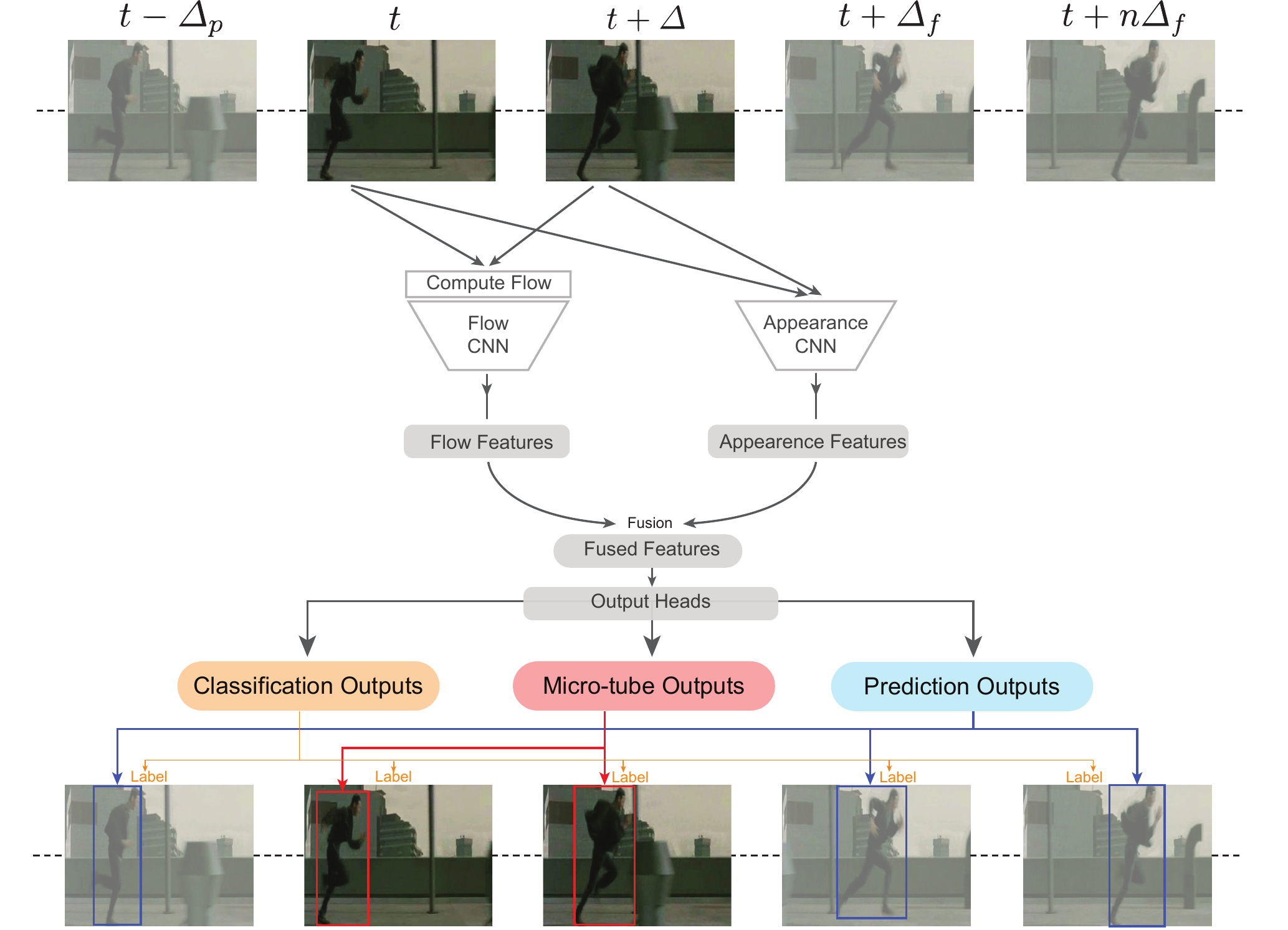}
  \vspace{-4mm}
  \caption{
    {\small
      \textit{Workflow illustrating the application of TPnet to a test video at a time instant $t$. 
      The network takes frames $f_t$ and $f_{t+\Delta}$ as input and 
      generates classification scores, the
      micro-tube (in red) for frames $f_t$ and $f_{t+\Delta}$, and
      prediction bounding boxes (in blue) for frames $f_{t-\Delta_p}$, $f_{t+\Delta_f}$ up to $f_{t+n\Delta_f}$. All bounding boxes are considered to be linked to the micro-tube.
      Note that predictions also span the past: a setting called \emph{smoothing} in the estimation literature. $\Delta_p$, $\Delta_f$ and $n$ are network parameters that we cross-validate during training.}
    }
 }
\label{fig:testing}
\vspace{-7mm}
\end{figure*}
\noindent
\textit{\textbf{Concept:}} We propose to extend the action micro-tube detection 
architecture by Saha \etal ~\cite{saha2017amtnet} to produce, at any time $t$, past ($\tau<t$), present, and future ($\tau>t$) detection bounding boxes, 
so that each (extended) micro-tube contains bounding boxes for both observed and not yet observed frames. All bounding boxes, spanning presently observed frames as well as past and future ones (in which case we call them predicted bounding boxes), are considered to be linked, as shown in blue in Figure~\ref{fig:testing}.
\\
We call this new deep network `TPnet'.
\\
Once bounding boxes are regressed, the online tube construction method of Singh \etal~\cite{singh2016online} can be incrementally applied to the observed part of the video to generate one or more `detected' action tubes at any time instant $t$.
\\
Further, in virtue of TPnet and online tube construction, 
the temporally linked micro-tubes forming each 
currently detected action tube 
(spanning the observed segment of the video) 
also contain past and future estimated bounding boxes. 
As these predicted boxes are implicitly linked to the micro-tubes 
which compose the presently detected tube, the problem of linking 
to the latter the future bounding boxes, leading to a whole action tube, 
is automatically addressed. 


The proposed approach provides two main benefits: 
i) future bounding box predictions are implicitly linked to the present action tubes;
ii) as the method relies only on two consecutive frames separated by a constant distance $\Delta$, 
it is efficient enough to be applicable to real-time scenarios. 

\textit{\textbf{Contributions:}} In Summary we present a Tube Predictor network (TPnet) which: 
\begin{itemize}
    \item given a partially observed video, can (early) predict video long action tubes in terms of both their classes and the constituting bounding boxes;
    \item demonstrates that training a network to make predictions also helps in improving action detection performance;
    \item demonstrates that feature-based fusion works better than late fusion in the context of spatiotemporal action detection. 
\end{itemize}

%% file: text/soa.tex
\section{Related work}
\emph{Early label prediction}. 
Early, online action label prediction has been studied using dynamic bag of words \cite{ryoo2011human}, 
structured SVMs \cite{hoai2014max}, 
hierarchical representations \cite{lan2014hierarchical}, 
LSTMs\cite{yeung2015every} and Fisher vectors \cite{de2016online}. 
Recently, Yeung~\etal~\cite{yeung2015every,yeungcvpr2016} have proposed a variant of 
long short-term memory (LSTM) deep networks for modelling 
these temporal relations via multiple input and output connections.
Kong \etal~\cite{kong2017deep}, instead, 
make use of variational auto-encoders to predict a representation for the whole video and use it to determine the action category for the whole video as early as possible. 
Probabilistic approaches based on Bayesian networks \cite{nazerfard2013using}, 
Conditional Random Fields \cite{CAD120koppula2013learning} or 
Gaussian processes \cite{jiang2014modeling} may help in activity anticipation.
However, inference in such generative approaches is often expensive. 
None of these methods address the full online label and
spatiotemporal location prediction setting considered here. 

\emph{Online Action Detection}. 
Soomro~\etal~\cite{Soomrocvpr2016} have recently proposed
an online method which can predict an action's label 
and detect its location by observing a relatively smaller portion of the entire video sequence.
They use segmentation to perform online detection via SVM models trained on fixed length segments of the training videos.
Similarly, Singh \etal~\cite{singh2016online} have extended online action detection to 
untrimmed videos with help of an online tube construction algorithm built on the top of frame-level action bounding box detections.
Similiarly, Behl \etal~\cite{behl2017incremental} solve online detection with help of tracking formulation.
However, these approaches~\cite{Soomrocvpr2016,singh2016online,behl2017incremental} only perform
action localisation for the observed part of the video and 
adopt the label predicted for the currently detected tube as the label for the whole video.

To the best of our knowledge, no existing method generates predictions concerning both labels and action tube geometry.  
Interestingly, Yang \etal \cite{yang2017spatio} use features from current,
frame $t$ proposals to `anticipate' region proposal locations in $t+\Delta$ 
and to generate detections at time $t+\Delta$, 
thus failing to take full advantage of 
the anticipation trick to predict the future spatiotemporal extent of the action tubes.

Advances in action recognition are always going to be helpful 
in action prediction from a general representation learning point of view.
For instance, Gu \etal \cite{ava2017gu} have recently improved on 
\cite{peng2016eccv,kalogeiton2017action} by plugging in the inflated 3D 
network proposed by \cite{carreira2017quo} as 
a base network on multiple frames.
Although they use a very strong base network pre-trained 
on the large ``Kinetics''~\cite{kay2017kinetics} dataset, 
they do not handle the linking process within the network as the 
AVA~\cite{ava2017gu} dataset's annotations are not temporally linked.  
Analogously, learning to predict future
representation~\cite{vondrick2015anticipating} can be useful in general action prediction (cfr. e.g.~\cite{kong2017deep}).

Recently, inspired by the record-breaking performance of 
CNN-based object detectors~\cite{redmon2016yolo9000,ren2015faster,liu15ssd}, 
a number of scholars~\cite{singh2016online,Saha2016,Georgia-2015a,peng2016eccv,Weinzaepfel-2015,weinzaepfel2016towards,zolfaghari2017chained,behl2017incremental} 
have tried to extend frame-level object detectors to videos 
for spatio-temporal action localisation.
These approaches, however, fail to tackle spatial and 
temporal reasoning jointly at the network level, 
as spatial detection and temporal association 
are treated as two disjoint problems.
More recent works have attempted to address this problem by reducing the amount 
of linking required with the help of `micro-tubes'~\cite{saha2017amtnet} 
or `tubelets'~\cite{kalogeiton2017action,hou2017tube} 
for small sets of frames taken together, where micro-tube boxes 
from different frames are considered to be linked together. 
AMTnet~\cite{saha2017amtnet} by Saha \etal \; is particularly interesting,
because of its compact (GPU memory-wise) and flexible 
nature, as it can exploit pairs of successive frames $\Delta$ sampling intervals apart, 
that it can also leverage sparse annotations~\cite{daly2016weinzaepfel} as well. 
For these reasons in this work we build on
AMTnet as base network, improving its feature representation by feature-level fusion of motion and appearance cues.

%% file: text/approach.tex
\section{Methodology}~\label{sec:methodology}

In this section, we describe our tube prediction framework for the problem formulation described in \S~\ref{subsec:problem_statement}.
Our approach has four main components.
Firstly, we tie the future action tube prediction problem (\S~\ref{subsec:problem_statement}) 
with action micro-tube~\cite{saha2017amtnet} detection. 
Secondly, we devise our tube prediction network (TPnet) 
to predict future bounding boxes along with current micro-tubes,
and describe its training procedure in~\S~\ref{subsec:training}.
Thirdly, we use TPnet in a sliding window fashion (\S~\ref{subsec:predict}) 
in the temporal direction while generating micro-tubes and corresponding future predictions. 
These, eventually, are fed to a tube prediction framework (\S~\ref{subsec:predict}) 
to generate the future of any current action tube being built using micro-tubes.

\subsection{Problem Statement}\label{subsec:problem_statement}
We define an \emph{action tube} as a connected sequence of detection boxes 
in time without interruptions and associated with a same action class $c$, 
starting at first frame $f_1$ and ending last frame $f_T$, 
in trimmed video: $\mathcal{T}_c = \{ {b}_{1}, ... {b}_{t}, ... {b}_{T}\}$.
Tubes are constrained to span the entire video duration, like in ~\cite{Georgia-2015a}.
At any time point $t$, a tube is divided into two parts, one needs 
to be detected $\mathcal{T}_{c}^{d} = \{ {b}_{1}, ... {b}_{t}\}$ up to $f_t$ and 
another part needs to be predicted/estimated
$\mathcal{T}_{c}^{p} = \{{b}_{t+1}, ... {b}_{T}\}$ from frame $f_{t+1}$ to $f_{T}$ 
along with its class $c$.
The observed part of the video is responsible for generating 
$\mathcal{T}_{c}^{d}$ (red in Fig~\ref{fig:problem-statement}), 
while we need to estimate the future section of the tube $\mathcal{T}_{c}^{p}$ 
(blue in Fig~\ref{fig:problem-statement}) for the unobserved segment of the video.
The first sub-problem, the online detection of $\mathcal{T}_{c}^{d}$, 
is explained in \S~\ref{subsec:micro-tube}. 
The second sub-problem (the estimation of the future tube segment $\mathcal{T}_{c}^{p}$) 
is tackled by a tube prediction network (TPnet, \S~\ref{subsec:training}) 
in a novel tube prediction framework (\S~\ref{subsec:predict}).

\subsection{From micro-tubes to full action tubes} \label{subsec:micro-tube}

\begin{figure*}[t]
  \centering
  \includegraphics[scale=0.23]{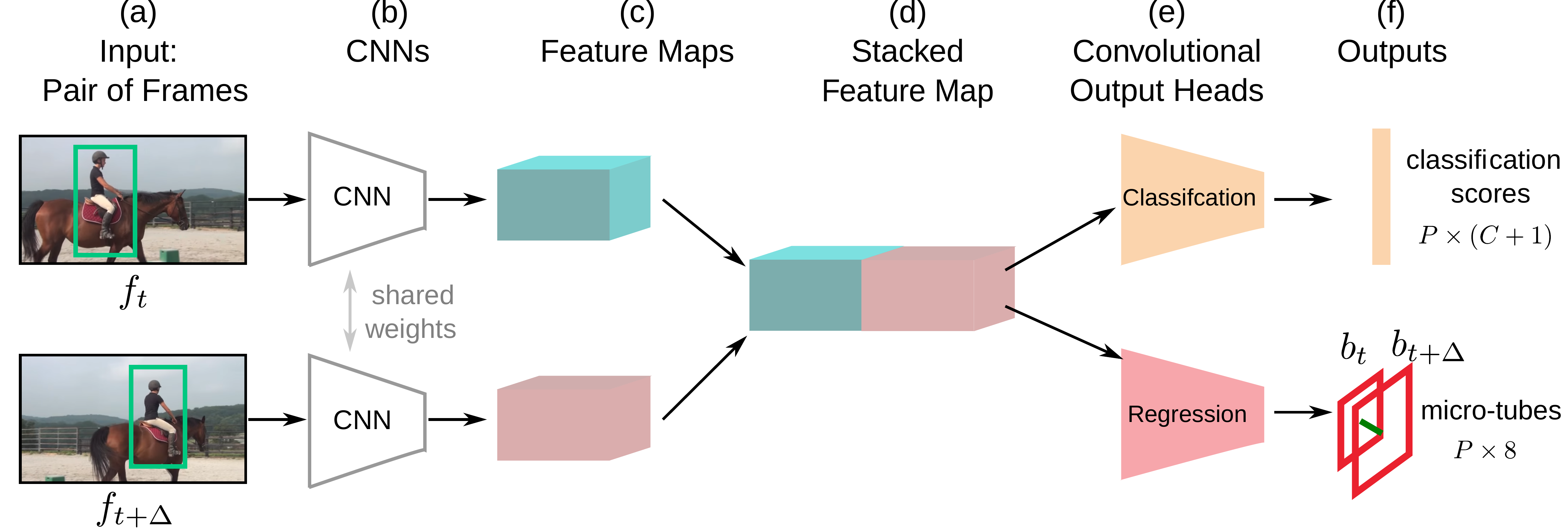}
  \caption{
    {\small
      \textit{Overview of the action micro-tube detection network (AMTnet). As it only predicts micro-tubes and their scores, here we modify it to predict the future locations associated with the given micro-tubes, as shown in Figure~\ref{fig:testing}.}
    }
 }
\label{fig:amtnet}
\end{figure*}

Saha \etal~\cite{saha2017amtnet} introduced \emph{micro-tubes} in their action micro-tube network (AMTnet) proposal, shown in Figure~\ref{fig:amtnet}.
AMTnet decomposes the problem of detecting $\mathcal{T}_c$ into a set of smaller problems, detecting micro-tubes ${m}_t = \{b_t,b_{t+\Delta}\}$ at time $t$ along with their classification scores for $C+1$ classes, using two successive frames $f_t$ and $f_{t+\Delta}$ as an input (Fig.~\ref{fig:amtnet}(a)). Subsequently, the detection micro-tubes $\{{m}_{1} ... {m}_{t-\Delta}\}$ are linked up in time to form action tube $\mathcal{T}_c^d$.
Similar to~\cite{liu15ssd}, one background class is added to the class list which takes the number classes to $C+1$.

AMTnet employs two parallel CNN streams (Fig.~\ref{fig:amtnet}(b)), one for each frame, 
which produce two feature maps (Fig.~\ref{fig:amtnet}(c)).
These feature maps are stacked together into one (Fig~\ref{fig:amtnet}(d)).
Finally, convolutional heads are applied in a sliding window (spatial) fashion over
predefined $3\times3$ anchor regions~\cite{liu15ssd}, which correspond
to $P$ prior~\cite{liu15ssd} or anchor~\cite{ren2015faster} boxes.
Convolutional heads produce a $P\times8$ output per micro-tube (Fig.~\ref{fig:amtnet}(f)) 
and $P\times(C+1)$ corresponding classification scores (Fig ~\ref{fig:amtnet}(g)).
Each micro tube has $8$ coordinate, $4$ for the bounding box $b_t$ in frame $f_t$ 
and $4$ for bounding box $b_{t+\Delta}$ in frame $f_{t+\Delta}$. 
As shown in Figure~\ref{fig:amtnet}(f), the pair of boxes can be considered as implicitly linked together, hence the name micro-tube.

Originally, Saha \etal~\cite{saha2017amtnet} employed FasterRCNN~\cite{ren2015faster} 
as base detection architecture. Here, however, we switch to Single Shot Detector (SSD) \cite{liu15ssd} as 
a base detector for efficiency reasons.
Singh \etal~\cite{singh2016online} used SSD to propose an online and real-time action tube generation algorithm, while
Kalogeiton \etal~\cite{kalogeiton2017action} adapted SSD 
to detect micro-tubes (or, in their terminology, `tubelets') $k$ frames long.
\\
More importantly, we make two essential changes to AMTnet. 
Firstly, we enhance its feature representation power by 
fusing appearance features (based on RGB frames) and flow features (based on optical flow) 
at the feature level (see the fusion step shown in Fig~\ref{fig:training}), 
unlike the late fusion approach of \cite{kalogeiton2017action} and \cite{singh2016online}. 
Note that the original AMTnet framework does not make use of optical flow at all. 
We will show that feature-level fusion dramatically improves its performance.
Secondly, the AMTnet-based tube detection framework proposed in \cite{saha2017amtnet} 
is offline, as micro-tube linking is done recursively in an offline fashion \cite{Georgia-2015a}.
Similar to Kalogeiton \etal~\cite{kalogeiton2017action}, 
we adapt the online linking method of \cite{singh2016online} 
to link micro-tubes in to a tube $\mathcal{T}_{c}^{d}$.

\noindent
\textbf{Micro-tube linking details:} 
Let $B_t$ be the set of detection bounding boxes from 
frame $f_t$, and $B_{t+1}$ the corresponding set from $f_{t+1}$, generated by a frame-level detector.
Singh \etal~\cite{singh2016online} associate boxes in $B_t$ to boxes in $B_{t+1}$,
whereas, in our case, we need to link
micro-tubes $m_t \in M_t \doteq B_t^{1} \times B_{t+\Delta}^{2}$ 
from a pair of frames $\{f_t,f_{t+\Delta}\}$ to microtubes
$m_{t+\Delta} \in M_{t+\Delta} \doteq B_{t+\Delta}^1 \times B_{t+2\Delta}^2$ from 
the next set of frames 
$\{f_{t+\Delta},f_{t+2\Delta}\}$.
This happens by associating elements of $B_{t+\Delta}^2$, coming
from $M_t$, with elements of $B_{t+\Delta}^1$, coming from $M_{t+\Delta}$.
Interestingly, the latter is a relatively easier sub-problem, 
as all such detections are generated based on the same frame, 
unlike the across frame association problem considered in ~\cite{singh2016online}. 
The association is achieved based on Intersection over Union (IoU) and class score, as 
the tubes are built separately for each class in a multi-label scenario.
For more details, we refer the reader to~\cite{singh2016online}.

\begin{figure*}[t]
  \centering
  \includegraphics[scale=0.55]{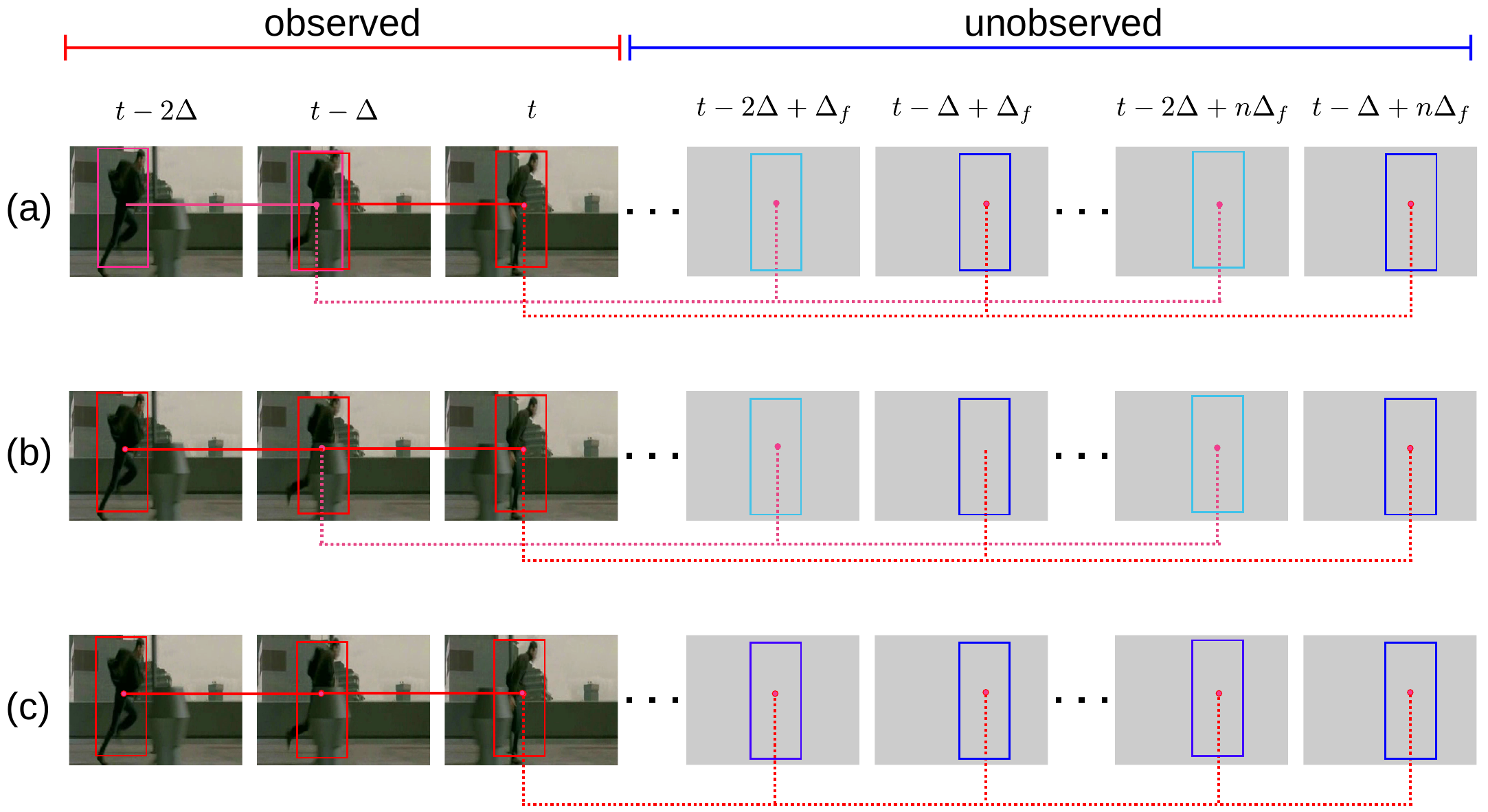}
  \caption{
    {\small
      \textit{Overview of future tube ($\mathcal{T}_{c}^{p}$)
      prediction using the predictions that are linked to micro-tubes.
      The first row (a) shows two output micro-tubes in light red and red and 
      their corresponding predictions in future in light blue and blue.
      In row (b) two micro-tubes are linked together, after which 
      they are shown in the same colour (red).
      By induction on the previous step, in row (c) we show that the predictions associated with two micro-tubes are linked together as well, hence forming one single tube. The observed segment is shown in red, while the predicted segment for the part of the video yet to observe
      is shown in blue.}
    }
 }
\label{fig:linking}
\end{figure*}

Since we adopt the online linking framework of Singh \etal~\cite{singh2016online}, 
we follow most of the linking setting used by them, e.g.: 
linking is done for every class separately; the non-maximal threshold is set to $0.45$. 
As shown in Figure~\ref{fig:linking}(a) to \ref{fig:linking}(b), 
the last box of the first micro-tube (red) is linked to the first box of next micro-tube (red). 
So, the first set of micro-tubes is produced at $f_1$, the following one at $f_{\Delta}$ 
the one after that at $f_{2\Delta}$, and so on. As a result, 
the last micro-tube is generated at $f_{t-\Delta}$ to cover 
the observable video duration up to time $t$.
Finally, we solve for the association problem as described above.


\subsection{Training the tube prediction network (TPnet)}\label{subsec:training}
AMTnet allow us to detect current tubes $\mathcal{T}_{c}^{d}$ 
by generating a set of successive micro-tubes 
$\{{m}_{1} ... {m}_{t-\Delta}\}$, where ${m}_{t-\Delta} = \{b_{t-\Delta},b_t\}$.
However, our aim is to predict the future section $\mathcal{T}_{c}^{p}$ of the tube 
using the latter linked micro-tubes, up to time $t$.

To address this problem, we propose a tube prediction framework aimed at
simultaneously estimating a micro-tube ${m}_t$, a set 
${z}_t = \{{b_{t-\Delta_p}, 
b_{t+\Delta_f}, ... b_{t+n\Delta_f}}\}$ 
of \emph{past and future detections}, 
and the classification scores for the $C+1$ classes.
$\Delta_p$ measures how far in the past we are looking into,
whereas $\Delta_f$ is a future step size, and $n$ is the number of future steps.
This is performed by a new Tube Prediction network (TPnet).

\begin{figure*}[t]
  \centering
  \includegraphics[scale=0.60]{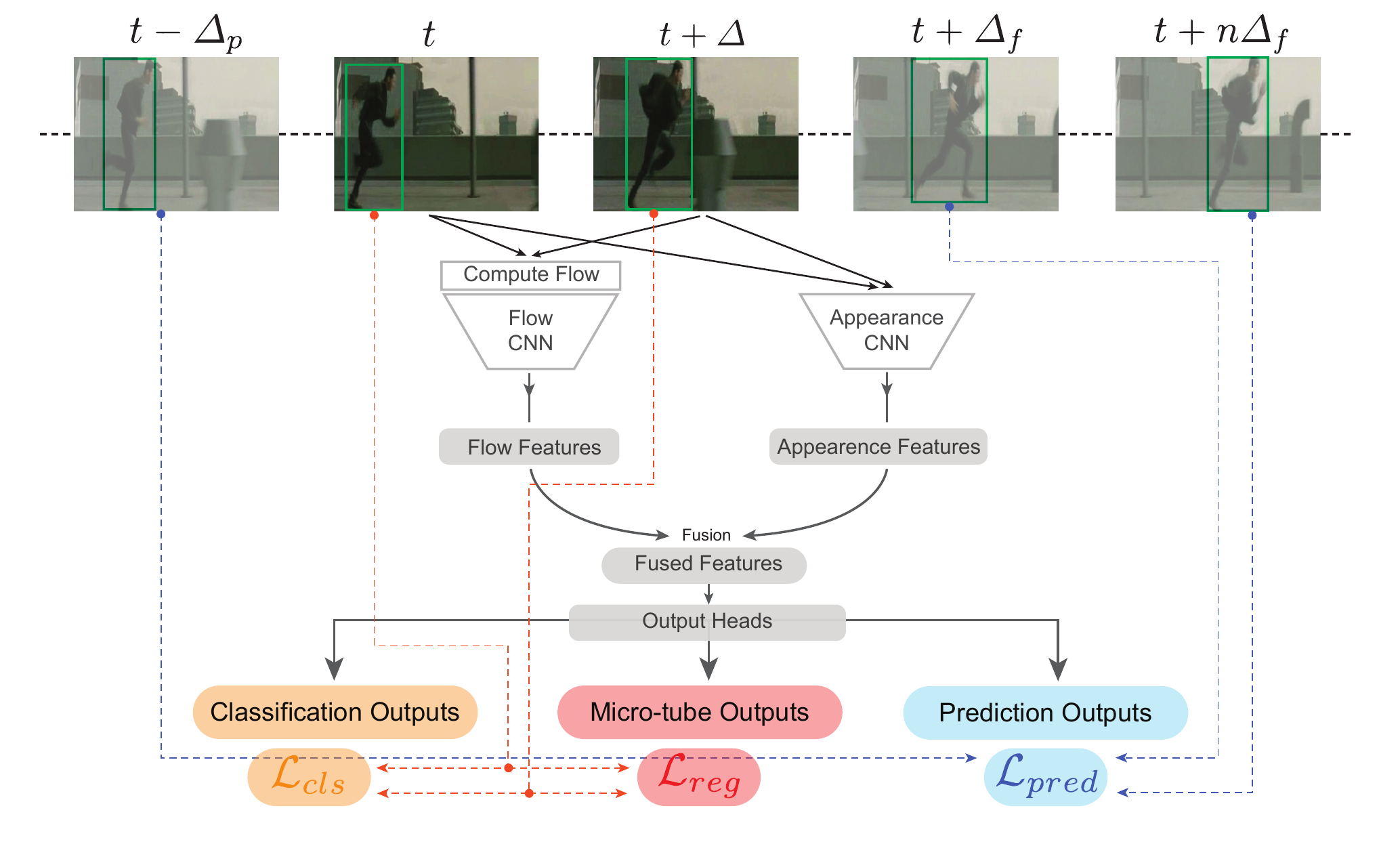}
  \caption{
    {\small
      \textit{Overview of the tube prediction network (TPnet) architecture at training time.}
    }
 }
\label{fig:training}
\end{figure*}

The underlying architecture of TPnet is shown in Figure~\ref{fig:training}.
TPnet takes two successive frames from time $t$ and $t+\Delta$ as input.
The two input frames are fed to two parallel CNN streams, 
one for appearance and one for optical flow.
The resulting feature maps are fused together, either by concatenating or by element-wise summing 
the given feature maps.
Finally, three types of convolutional output heads are used for 
$P$ prior boxes as shown in Figure~\ref{fig:training}.
The first one produces the $P\times(C+1)$ classification outputs; the
second one regresses the $P\times 8$ coordinates of the micro-tubes, as in AMTnet; the 
last one regresses $P\times(4(1+n))$ coordinates, 
where $4$ coordinates correspond to the frame at $t-\Delta_p$, and the remaining $4n$ are associated with the $n$ future steps. 
The training procedure of the new architecture is illustrated below.

\noindent
\textbf{Multi-task learning}\label{sss:multitask}
TPnet is designed to strive for three objectives, for each prior box $p$. 
The first task (i) is to classify the $P$ prior boxes; 
the second task (ii) is to regress the coordinates of the micro-tubes;  
the last (iii) is to regress the coordinates of the past and future detections associated with each micro-tube.

Given a set of $P$ anchor boxes and the respective outputs 
we compute a loss following 
the training objective of SSD~\cite{liu15ssd}. Let $x_{i,j}^c = \{0,1\}$ be the indicator  
for matching the $i$-th prior box to the $j$-th ground truth box of category $c$. 
We use the bipartite matching procedure described in~\cite{liu15ssd} 
for matching the ground truth micro-tubes $G = \{g_t,g_{t+\Delta}\}$ 
to the prior boxes, where $g_t$ is a ground truth box at time $t$.
The overlap is computed between a prior box $p$ and micro-tube $G$ as the 
mean IoU between $p$ and the ground truth boxes in $G$.
A match is defined as positive ($x_{i,j}^c = 1$) if the overlap is more than or equal to $0.5$.  

The overall loss function $\mathcal{L}$ is 
the following weighted sum of classification loss ($\mathcal{L}_{cls}$), 
micro-tube regression loss ($\mathcal{L}_{reg}$) and prediction loss ($\mathcal{L}_{pred}$):
\begin{equation}
\mathcal{L}(x,c,m,G,z,Y) = \frac{1}{N} \big ( \mathcal{L}_{cls}(x,c) + \alpha\mathcal{L}_{reg}(x,m,G) + \beta\mathcal{L}_{pred}(x,z,Y) \big ),
\end{equation}
where $N$ is the number of matches, $c$ is the ground truth class, 
$m$ is the predicted micro-tube, $G$ is the ground truth micro-tube,
$z$ assembles the predictions for the future and the past, and $Y$ is the ground truth of future and past bounding boxes associated with the ground truth micro-tube $G$. 
The values of $\alpha$ and $\beta$ are both set to $1$ in all of our experiments:
different values might result in better performance.

The classification loss $\mathcal{L}_{cls}$ is a softmax cross-entropy loss; 
a hard negative mining strategy is also employed, as proposed in~\cite{liu15ssd}.
The micro-tube loss $\mathcal{L}_{reg}$ is a Smooth L1 loss~\cite{ren2015faster} 
between the predicted ($m$) 
and the ground truth ($G$) micro-tube. 
Similarly, the prediction loss $\mathcal{L}_{pred}$ is also a Smooth L1 loss between the predicted boxes ($z$) and the ground truth boxes ($Y$). 
As in \cite{liu15ssd,ren2015faster}, 
we regress the offsets with respect to the coordinates of 
matched prior box $p$ matched to $G$ for both $m$ and $z$.
We use the same offset encoding scheme as used in~\cite{liu15ssd}. 

\subsection{Tube prediction framework}\label{subsec:predict}

TPnet is shown in Figure~\ref{fig:testing} at test time. As in the training setting,
it observes only two frames 
that are $\Delta$ apart at any time point $t$. 
The outputs of TPnet at any time $t$ are linked to a micro-tube,
each micro-tube containing a set of bounding boxes 
$\{m_t = \{b_t, b_{t+\Delta}\}; z_t = \{b_{t-\Delta_p},b_{t+\Delta_f},...,b_{t+\Delta_f}\}\}$, 
which are considered as linked together.

As explained in \S~\ref{subsec:micro-tube}, given a set of 
micro-tubes $\{{m}_{1} ... {m}_{t-\Delta}\}$ 
we can construct $\mathcal{T}_{c}^{d}$ by online linking~\cite{singh2016online}
of the micro-tubes. As a result, we can use predictions for $t+\Delta_f$ up to $t+n\Delta_f$ to generate the future of $\mathcal{T}_{c}^{d}$, thus extending it further into the future
as shown in Figure~\ref{fig:linking}. More specifically,  
as it is indicated in Figure~\ref{fig:linking}(a), 
a micro tube at $t-2\Delta$ is composed by $n+2$ bounding 
boxes ($\{b_{t-2\Delta},b_{t-\Delta}, b_{t-2\Delta+\Delta_f}, ... b_{t-\Delta+n\Delta_f}\}$) linked together.
The last micro-tube is generated from $t-\Delta$.
In the same fashion, putting together the predictions associated with
all the past micro-tubes ($\{{m}_{1} ... {m}_{t-\Delta}\}$) 
yields a set of linked future bounding boxes 
($\{b_{t+1}, ... , b_{t+\Delta+\Delta_f}, ... , b_{t-\Delta+n\Delta_f}\}$) 
for the current action tube $\mathcal{T}_{c}^{d}$, thus outputting a part of the desired future $\mathcal{T}_{c}^{p}$.

Now, we can generate future tube $\mathcal{T}_{c}^{p}$ from 
the set of linked future bounding boxes 
($\{b_{t+1}, ... b_{t-\Delta+\Delta_f},... b_{t-\Delta+n\Delta_f}\}$)
from $t+1$ to $t-\Delta+n\Delta_f$ and 
simple linear extrapolation of 
bounding boxes from $t-\Delta + n\Delta_f$ to $T$.
Linear extrapolation is performed based on the average velocity 
of the each coordinates from last $5$ frames, 
predictions outside the image coordinate are trimmed to the image edges.

%% file: text/experiments_results_discussion.tex
\section{Experiments} \label{sec:experiments}
We test our action tube prediction framework (\S~\ref{sec:methodology}) 
on four challenging problems: 
i) action localisation (\S~\ref{sec:localisation}),
ii) early action prediction (\S~\ref{sec:prediction-online-localisation}),
iii) online action localisation (\S~\ref{sec:prediction-online-localisation}),
iv) future action tube prediction (\S~\ref{sec:future-localisation})
Finally, evidence of real time capability is quantitatively demonstrated in (\S~\ref{sec:detection-speed}).

\textbf{J-HMDB-21.}
We evaluate our model on the J-HMDB-21~\cite{J-HMDB-Jhuang-2013} benchmark.
J-HMDB-21~\cite{J-HMDB-Jhuang-2013}  
is a subset of the HMDB-51 dataset~\cite{HMDBkuehne2011hmdb} 
with 21 action categories and 928 videos, 
each containing a single action instance 
and trimmed to the action's duration.
It contains atomic action which are 20-40 frames long.
Although, videos are of short duration (max $40$ frames), 
we consider this dataset because tubes belong to the same class and 
we think it is a good dataset to start with for action prediction task.

\textbf{Evaluation metrics.}
Now, we define the evaluation metrics used in this paper. 
i) We use a standard mean-average precision metric to evaluate the detection performance when the whole video is observed.
ii) Early label prediction task is evaluated by 
video classification accuracy~\cite{Soomrocvpr2016,singh2016online} 
as early as when only 10\% of the video frames are observed. 
\\
iii) Online action localisation (\S~\ref{sec:prediction-online-localisation}) is set up based 
on the experimental setup of~\cite{singh2016online}, 
and use mAP (mean average precision) as metric for online action detection 
i.e. it evaluates present tube ($\mathcal{T}_{c}^{d}$) built in online fashion.
\\
iv) The future tube prediction is a new task; we propose to evaluate its performance in two ways.
Firstly, we evaluate the quality of the whole tube prediction from the start of the videos to the end as early as when only 10\% of the video is observed. 
The entire tube predicted (by observing only a small portion (\%) of the video) is compared against the ground truth tube for the whole video.
Based on the detection threshold we can compute mean-average-precision for the complete tubes, we call this metric \emph{completion-mAP} (c-mAP).
Secondly, we measure how well the future predicted part of the tube localises.
In this measure, we compare the predicted ($\mathcal{T}_{c}^{p}$) tube with the corresponding ground truth future tube segment. 
Given the ground truth and the predicted future tubes, we can compute the mean-average precision for the predicted tubes, we call this metric \emph{prediction-mAP} (p-mAP).

We report the performance of previous three tasks (i.e. task ii to iv) as a function of \emph{Video Observation Percentage}, i.e., the portion (\%) of the entire video observed. 

\noindent
\textbf{Baseline.}
We modified AMTnet to fuse flow and appearance features~\ref{subsec:micro-tube}. 
We treat it as a baseline for all of our tasks.
Firstly, we show how feature fusion helps AMTnet in Section~\ref{sec:localisation}, 
and compare it with other action detection methods along with our TPnet.
Secondly in Section~\ref{sec:future-localisation}, 
we linearly extrapolate the detection from AMTnet to construct the future tubes, 
and use it as a baseline for tube prediction task. 
\noindent
\textbf{Implementation details.}
We train all of our networks with the same set of hyper-parameters 
to ensure the fair comparison and consistency, including TPnet and AMTnet.
We use an initial learning rate of $0.0005$, 
and the learning rate drops by a factor of $10$ after $5K$ and $7K$  iterations.
All the networks are trained up to $10K$ iterations. 
We implemented AMTnet using pytorch (\url{https://pytorch.org/}).
We initialise AMTnet and TPnet models using the pretrained SSD network on J-HMDB-21 dataset on its respective train splits. 
The SSD network training is initialised using image-net trained VGG network.
For, optical flow images, we used optical flow algorithm of Brox \etal~\cite{Brox-2004}.
Optical flow output is put into a three channel image, two channels are made of flow vector and the third channel is the magnitude of the flow vector. 

\begin{table}[t]
  \centering
  \setlength{\tabcolsep}{4pt}
  \caption{Action localisation results on JHMDB dataset.
  The table is divided into four parts. 
  The first part lists approaches which takes a single frame as input;
  the second part presents approaches which takes multiple frames as input;
  the third part contemplates different fusion strategies of our feature-level 
  fusion (based on AMTnet);
  lastly, we report the detection performance of our TPnet by ignoring the future and past predictions and only use the detected micro-tubes to produce the final action tubes.}
  \vspace{-2mm}
  {\footnotesize
  \scalebox{1}{
  \begin{tabular}{lccccc}
  \toprule
  Methods & $\delta$ = 0.2 &$\delta$ = 0.5 & $\delta$ = 0.75 & $\delta$ = .5:.95 & Acc \%\\\midrule
  MR-TS Peng~\etal~\cite{peng2016eccv} & 74.1  & 73.1 & -- & -- & --\\
  FasterRCNN Saha~\etal~\cite{Saha2016}   & 72.2  & 71.5 & 43.5 & 40.0 & --\\
  OJLA Behl~\etal~\cite{behl2017incremental}${}^*$  & -- & 67.3 & -- & 36.1 & --\\
  SSD Singh~\etal~\cite{singh2016online}${}^*$  & 73.8 & 72.0 & 44.5 & 41.6 & --\\
  \midrule
  AMTnet Saha~\etal~\cite{saha2017amtnet} rgb-only  & 57.7  & 55.3 & -- & -- & --\\
  ACT kalogeiton~\etal~\cite{kalogeiton2017action}${}^*$   & 74.2 & 73.7 & 52.1 & 44.8 & 61.7 \\
  T-CNN (offline) Hou~\etal~\cite{hou2017tube} & \textbf{78.4} & 76.9   & -- & -- & 67.2 \\
  MR-TS~\cite{peng2016eccv} + I3D~\cite{carreira2017quo} Gu \etal~\cite{ava2017gu}  
  & --  & \textbf{78.6} & -- & -- & --\\
  \midrule
  AMTnet-LateFusion${}^*$  & 71.7 & 71.2 & 49.7 & 42.5 & 65.8 \\
  AMTnet-FeatFusion-Concat${}^*$  & 73.1 & 72.6 & 59.8 & 48.3 & 68.4 \\
  AMTnet-FeatFusion-Sum${}^*$  & 73.5 & 72.8 & 59.7 & 48.1 & 69.6 \\
  \midrule
  Ours TPnet${}_{053}$${}^*$ & 72.6 & 72.1 & 58.0 & 46.7 & 67.5 \\
  Ours TPnet${}_{453}$${}^*$ & 73.8 & 73.0 & 59.1 & 47.3 & 68.2 \\
  Ours TPnet${}_{051}$${}^*$ & 74.6 & 73.1 & 60.5 & 49.0 & \textbf{69.8} \\
  Ours TPnet${}_{451}$${}^*$ & 74.8 & 74.1 & \textbf{61.3} & \textbf{49.1} & 68.9 \\ \bottomrule
  \multicolumn{6}{l}{ TPnet${}_{abc}$ represents our TPnet where $a = \Delta_p$, $b = \Delta_f$ and $c = n$.; ${}^*$ means online methods}\\
  \end{tabular}
  }
  }
  \label{table:localisation} \vspace{-6mm}
\end{table}

\noindent
\textbf{TPnet${}_{abc}$.} 
The training parameters of our TPnet are used to 
define the name of the setting in which we use our tube prediction network. 
The network name TPnet${}_{abc}$ represents our 
TPnet where $a = \Delta_p$, $b = \Delta_f$ and $c = n$, 
if $\Delta_p$ is set to $0$ it means network doesn't learn 
to predict the past bounding boxes.
In all of our settings, we use $\Delta=1$.

\subsection{Action localisation performance}\label{sec:localisation}
Table~\ref{table:localisation} shows the traditional action localisation results for 
the whole action tube detection in the videos of J-HMBD-21 dataset.
\\
\textbf{Feature fusion} compared to the late fusion scheme in AMTnet shows (Table~\ref{table:localisation}) remarkable improvement, at detection threshold $\delta=0.75$ the gain with feature level fusion is $10\%$, as a result, it is able to surpass the performance of ACT~\cite{kalogeiton2017action}, which relies on set of $6$ frames as compared to AMTnet which uses only $2$ successive frames as input.
Looking at the average-mAP ($\delta=0.5:95$), we can see that the fused model improves by almost $8\%$  as compared to single frame SSD model of Singh~\etal~\cite{singh2016online}.
We can see that concatenation and sum fusion perform almost similar for AMTnet.
Sum fusion is little less memory intensive on the GPUs 
as compared to the concatenation fusion; as a result, 
we use sum fusion in our TPnet.
\\
\textbf{TPnet for detection} is shown in the last part of the Table~\ref{table:localisation}, 
where we only use the detected micro-tubes by 
TPnet to construct the action tubes(\S~\ref{subsec:micro-tube}).
We train TPnet to predict future and past 
(i.e. when $\Delta_p>0$) as well as present micro-tubes.
We think that predicting bounding boxes for both the past 
and future video segments acts as a regulariser and helps 
improving the representation of the whole network.
Thus, improving the detection performance 
(Table~\ref{table:localisation}~TPNet${}_{051}$ and TPNet${}_{451}$).
However, that does not mean adding extra prediction task always 
help when a network is asked to learn prediction in far future, 
as is the case in TPNet${}_{053}$ and TPNet${}_{453}$,
we have a drop in the detection performance. 
We think there might be two possible reasons for this,  
i) network might starts to focus more on prediction task, and
ii) videos in J-HMDB-21 are short and number of training samples 
decreases drastically ($19K$ for TPNet${}_{051}$ and $10K$ for TPNet${}_{453}$), 
because we can not use edge frames of the video in training samples as 
we need a ground truth bounding box which is $15$ frames in the future, 
as $\Delta_f=5$ and $n=3$ for TPNet${}_{053}$.
However, in Section~\ref{sec:future-localisation}, 
we show that the TPNet${}_{053}$ model is the best to predict the future very early. 

\begin{figure*}[t]
  \centering
  \includegraphics[scale=0.38]{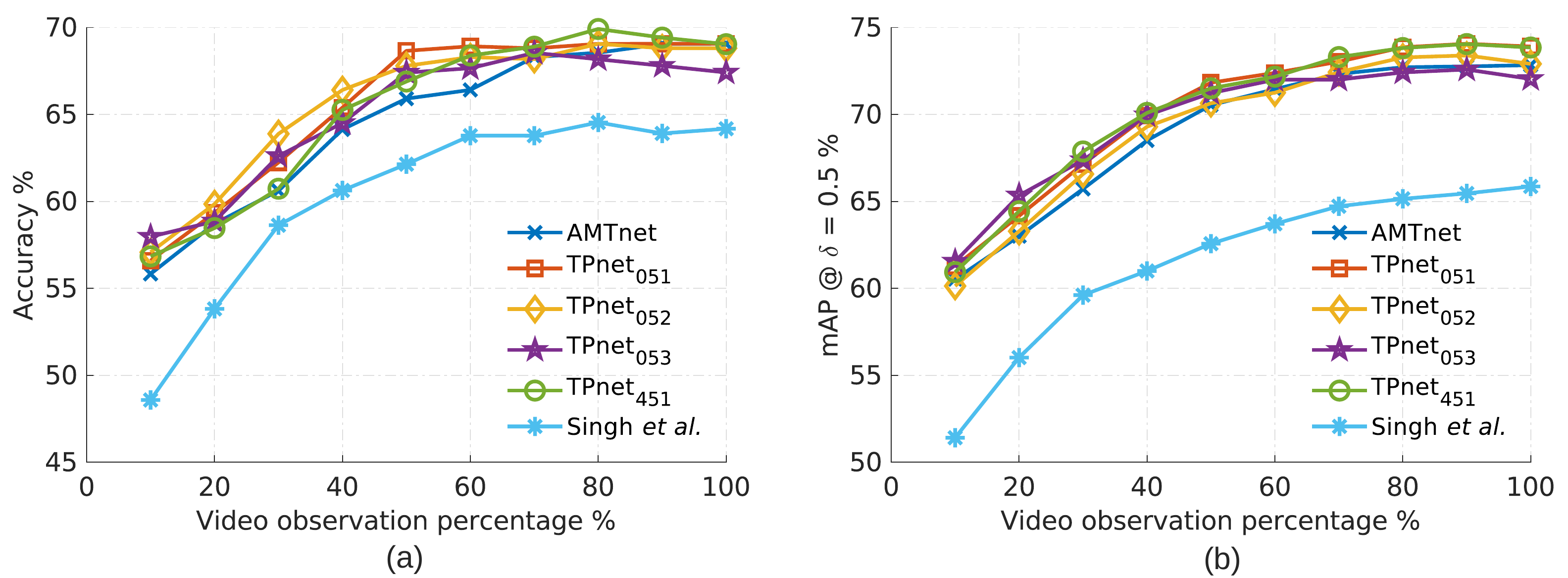}
  \vspace{-3mm}
  \caption{
    {\small
      \textit{Early label prediction results (video-level label prediction accuracy) 
      on J-HMDB-21 dataset in sub-figure (a). 
      Online action detection results (mAP with detection threshold $\delta = 0.5$) 
      on J-HMDB-21 dataset are shown in sub-figure (b).
      TPnet${}_{abc}$ represents our TPnet where $a = \Delta_p$, $b = \Delta_f$ and $c = n$.}
    }
 }
\label{fig:onlineNlabel}
\vspace{-5mm}
\end{figure*}

\subsection{Early label prediction and online localisation}\label{sec:prediction-online-localisation}
Figure~\ref{fig:onlineNlabel} (a) \& (b) show the early prediction 
and online detection capabilities of Singh~\etal~\cite{singh2016online}, 
AMTnet-Feature Fusion-sum and our TPnet.
\\
Soomro~\etal~\cite{Soomrocvpr2016}'s method also perform early label prediction on J-HMDB-21; however, their performance is deficient, as a result the plot would become skewed (Figure~\ref{fig:onlineNlabel}(a)), so we omit theirs from the figure. 
For instance, by observing only the initial $10\%$ of the videos in J-HMDB-21,
TPnet${}_{453}$ able to achieve a prediction accuracy of $58\%$ 
as compared to $48\%$ by Singh~\etal~\cite{singh2016online} and $5\%$ by Soomro~\etal~\cite{Soomrocvpr2016}, 
which is in fact higher than the $43\%$ accuracy achieved by \cite{Soomrocvpr2016} 
after observing the \emph{entire} video. 
As more and more video observed, all the methods improve, but TPnet${}_{451}$ show the most gain, 
however, TPnet${}_{053}$ observed the least gain from all the TPnet settings shown.
Which is in-line with action localisation performance discussed in the previous section~\ref{sec:localisation}.
We can observe the similar trends in online action localisation performance shown in Figure~\ref{fig:onlineNlabel}(b). 
To reiterate, TPnet${}_{053}$ doesn't get to see the training samples from the end portion of the videos, 
as it needs a ground truth bounding box from $15$ frames ahead. 
So, the last frame it sees of any training video is $T-15$, 
which is almost half the length of the most extended video($40$ frames) in J-HMDB-21. 
This effect magnifies when online localisation performance measured at $\delta=0.75$, we provide the evidence of it in the supplementary material.
\begin{figure*}[t]
  \centering
  \includegraphics[scale=0.38]{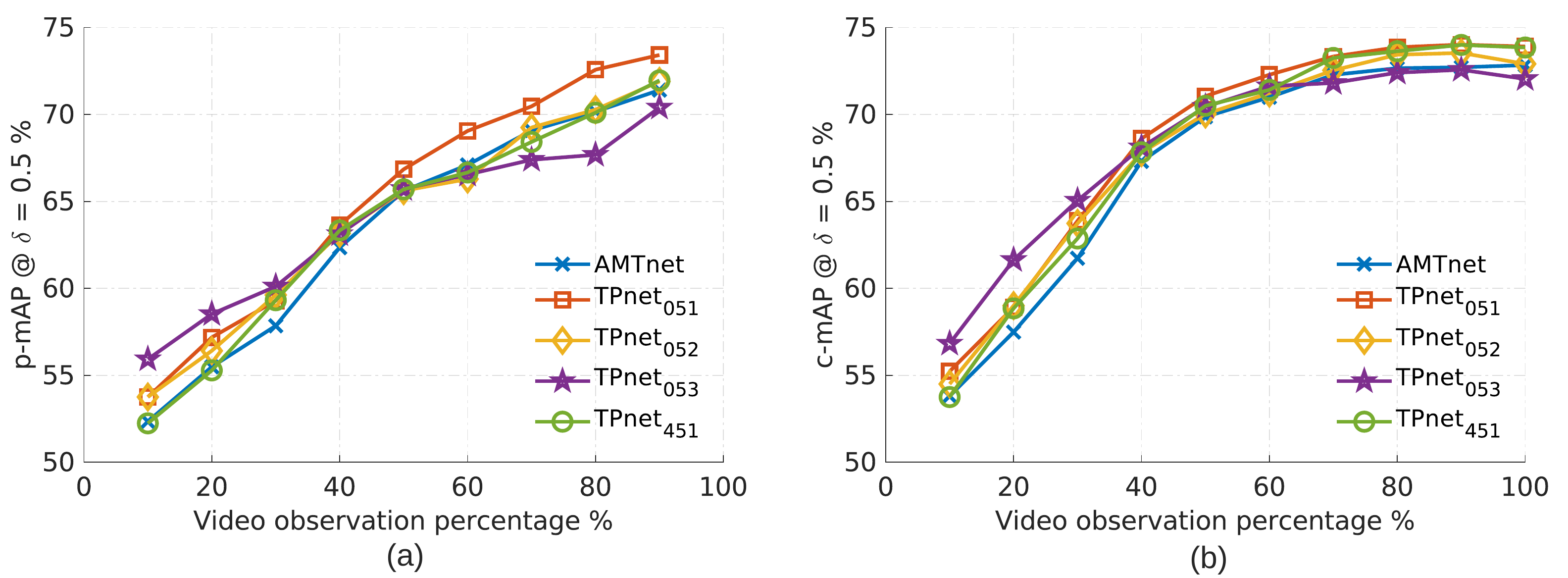}
  \vspace{1mm}
  \caption{
    {\small
      \textit{Future action tube prediction results (a) (prediction-mAP (p-mAP)) 
      for predicting the tube in unobserved part of the video.
      Action tube prediction results (b) (completion-mAP (c-mAP)) 
      for predicting video long tubes as early as possible 
      on J-HMDB-21 dataset in sub-figure (b).
      We use p-mAP (a) and c-mAP (b) with detection 
      threshold $\delta = 0.5$ as evaluation metrics on J-HMDB-21 dataset.
      TPnet${}_{abc}$ represents our TPnet where $a = \Delta_p$, $b = \Delta_f$ and $c = n$.}
    }
 }
\label{fig:prediction_mAP@05}
\end{figure*}

\subsection{Future action tube prediction}\label{sec:future-localisation}
Our main task of the paper is to predict the future of action tubes.
We evaluate it using two newly proposed metrics (\emph{p-mAP} and \emph{c-mAP}) 
as explained earlier at the start of the experiment section~\ref{sec:experiments}.
Result are shown in Figure~\ref{fig:prediction_mAP@05} for future tube prediction (Figure~\ref{fig:prediction_mAP@05} (a)) with p-mAP metric and 
tube completion with c-mAP as metric.

Although, the TPnet${}_{053}$ is the worst setting of TPnet model for early label prediction (Fig.~\ref{fig:onlineNlabel}(a)), online detection(Fig.~\ref{fig:onlineNlabel}(b)) and action tube detection (Table~\ref{table:localisation}), but as it predicts furthest in the future (i.e. $15$ frame away from the present time), 
it is the best model for early future tube prediction (Fig.~\ref{fig:prediction_mAP@05}(a)). 
However, it does not observe as much appreciation in performance as other settings as more and more frames are seen, owing to the reduction in the number of training samples. 
On the other hand, TPnet${}_{451}$ observed large improvement as 
compared to TPnet${}_{051}$ as more and more portion of the video is observed 
for tube completion task (Fig.\ref{fig:prediction_mAP@05}(b)), 
which strengthen our arguement that predicting not only the future but 
also the past is useful to achieve more regularised predictions.

\noindent
\textbf{Comparision with the baseline.}
As explained above, we use AMTnet as a baseline, and its results can be seen in all the plots and the Table. 
We can observe that our TPnet performs better than AMTnet in almost all the cases, 
especially in our desired task of early future prediction (Fig~\ref{fig:prediction_mAP@05}(a)) 
TPnet${}_{043}$ shows almost $4\%$ improvement in p-mAP (at $10\%$ video observation) over AMTnet.

\noindent
\textbf{Discussion.} 
Predicting further into the future is essential to produce any meaningful predictions (seen in TPnet${}_{053}$),  but at the same time, predicting past is helpful to improve overall tube completion performance. 
One of the reasons for such behaviour could be that J-HMDB-21 tubes are short (max $40$ frames long).
We think training samples for a combination of TPnet${}_{053}$ and TPnet${}_{451}$, i.e. TPnet${}_{453}$ are chosen uniformly over the whole video while taking care of absence of ground truth in the loss function could give us better of both settings. 
We show the result of TPnet${}_{453}$ in current training setting in supplementary material. 
The idea of regularising based on past prediction is similar to the one used by Ma~\etal~\cite{ma2016learning}.

\subsection{Test Time Detection Speed}
\label{sec:detection-speed}
Singh \etal ~\cite{singh2016online} showcase their method's online and real-time capabilities.
Here we use their online tube generation method for our tube prediction framework to inherit those properties.
The only question mark is TPnet's forward pass speed.
We thus measured the average time taken for 
a forward pass for a batch size of 1 as compared to 8 by ~\cite{singh2016online}.
A single forward pass takes 46.8 milliseconds to process one text example, 
showing that it can be run in almost real-time at 21fps with two streams on a single 1080Ti GPU.
One can improve speed even further by testing TPnet with $\Delta$ equal to $2$ or $4$ and obtain a speed improvement of  $2\times$ or $2\times$.
However, use of dense optical flow~\cite{Brox-2004}, which is slow, but as in ~\cite{singh2016online}, 
we can always switch to real-time optical~\cite{kroeger2016fast} with small drop in performance.

%% file: text/conclusion.tex
\section{Conclusions} \label{sec:conclusions}
We presented TPnet, a deep learning framework for future action tube prediction in videos which, unlike previous online tube detection methods~\cite{singh2016online,Soomrocvpr2016}, generates future of action tubes as early as when $10\%$ of the video is observed.
It can cope with the future uncertainty better than 
the baseline methods while remaining state-of-the-art in action detection task.
Hence, we provide a scalable platform to push the boundaries of action tube prediction research; 
it is implicitly scalable to multiple action tube instances 
in the video as future prediction is made for each action tube separately. 
We plan to scale TPnet for action prediction in temporally untrimmed videos in the future. 

%% file: text/supp_details.tex
In this supplementary material, we first present implementation details in Section~\ref{subsec:details}.
Next we show extra plots at different detection threshold, where we include TPnet${}_{453}$ 
as well, as mentioned in the main paper.

\section{Implementation details}\label{subsec:details}
All models are trained with batch size of $16$ on two 1080Ti GPUs (11GB VRAM each) for AMTnet and TPnet.
We used Pytorch library to implement all of the models (AMTnet, and TPnet) by following
the implementation of Singh \etal \cite{singh2016online} closely from their GitHub repository
\footnote{https://github.com/gurkirt/realtime-action-detection\label{https}}.

\textbf{Input Data Preparations.}
The number of input frames to CNNs is more than 1 in our case e.g. 2 in AMTnet. 
Optical-flow stream uses
a stack of 5 frames as input for each input frame in the sequence.
In our case, sequence length is more than (in case AMTnet and TPnet equal to 2),
so 2 stacks of 5 frames are used as input for flow stream in case of TPnet for flow stream.
Number of input frames for RGB stream is equal to sequence length e.g. 2 for TPnet.
Each optical flow image is 3 channel image, where first two channels are flow in  $x$ and $y$ direction respectively and
the third channel is magnitude (square-root of the sum of squares) of flow in both directions.
We computed dense optical flow between each pair of successive video frames using the algorithms of~\cite{Brox-2004}.

SSD relies on \textbf{data augmentation} to boost the performance,
we extended the data augmentation~\cite{liu15ssd} function provided by Singh \etal \cite{singh2016online}
in  their GitHub repository\footnotemark[\ref{https}]
to accept sequence of frames .
The same data augmentation was applied to all the frames in one input sequence as used in \cite{singh2016online,kalogeiton2017action} .

\textbf{VGG weight initialisation.}
Weights of VGG network (base network) are initialised with weights from a pre-trained ImageNet
model~\cite{liu2015parsenet}\footnote{\url{https://gist.github.com/weiliu89/2ed6e13bfd5b57cf81d6}}
for appearance- and flow-based SSD networks for both training the initial SSD model, 
similar to ~\cite{singh2016online}. 
Finally, weights of AMTnet and TPnet are initialised using 
trained SSD model on J-HMDB-21 dataset for both flow and appearance streams.
\clearpage
\section{Early Label Prediction Performance (Accuracy).}
\begin{figure*}[h]
\vspace{-8mm}
  \centering
  \includegraphics[scale=0.65]{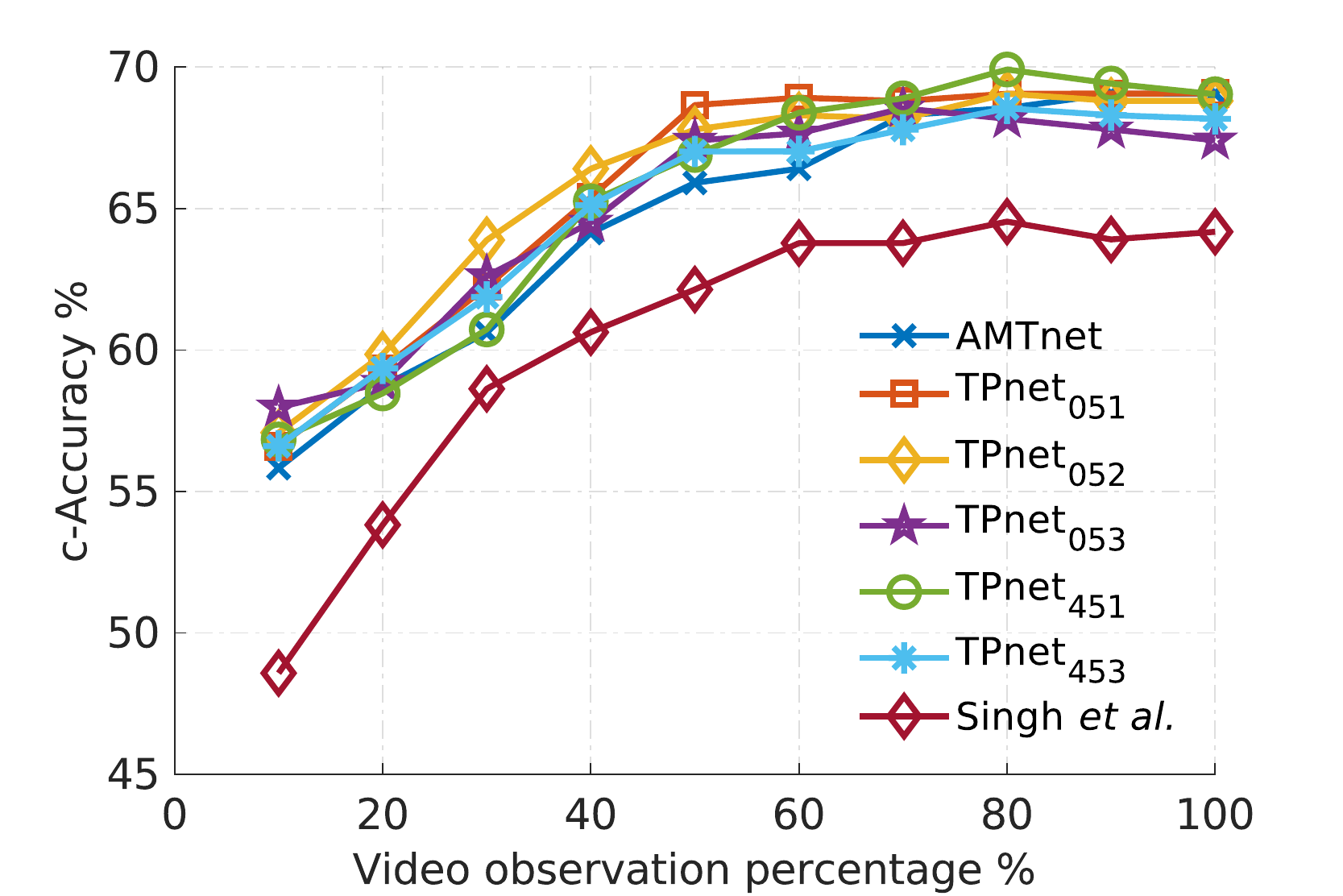}
  \caption{
    {\small
      \textit{Early label prediction results (video-level label prediction accuracy) 
      on J-HMDB-21 dataset in sub-figure.
      TPnet${}_{abc}$ represents our TPnet where $a = \Delta_p$, $b = \Delta_f$ and $c = n$.}
    }
 }
\label{fig:label}
\vspace{-10mm}
\end{figure*}


\clearpage
\section{Online Action Detection Performance (mAP).}
\begin{figure*}[h]
\vspace{-10mm}
  \centering
  \includegraphics[scale=0.60]{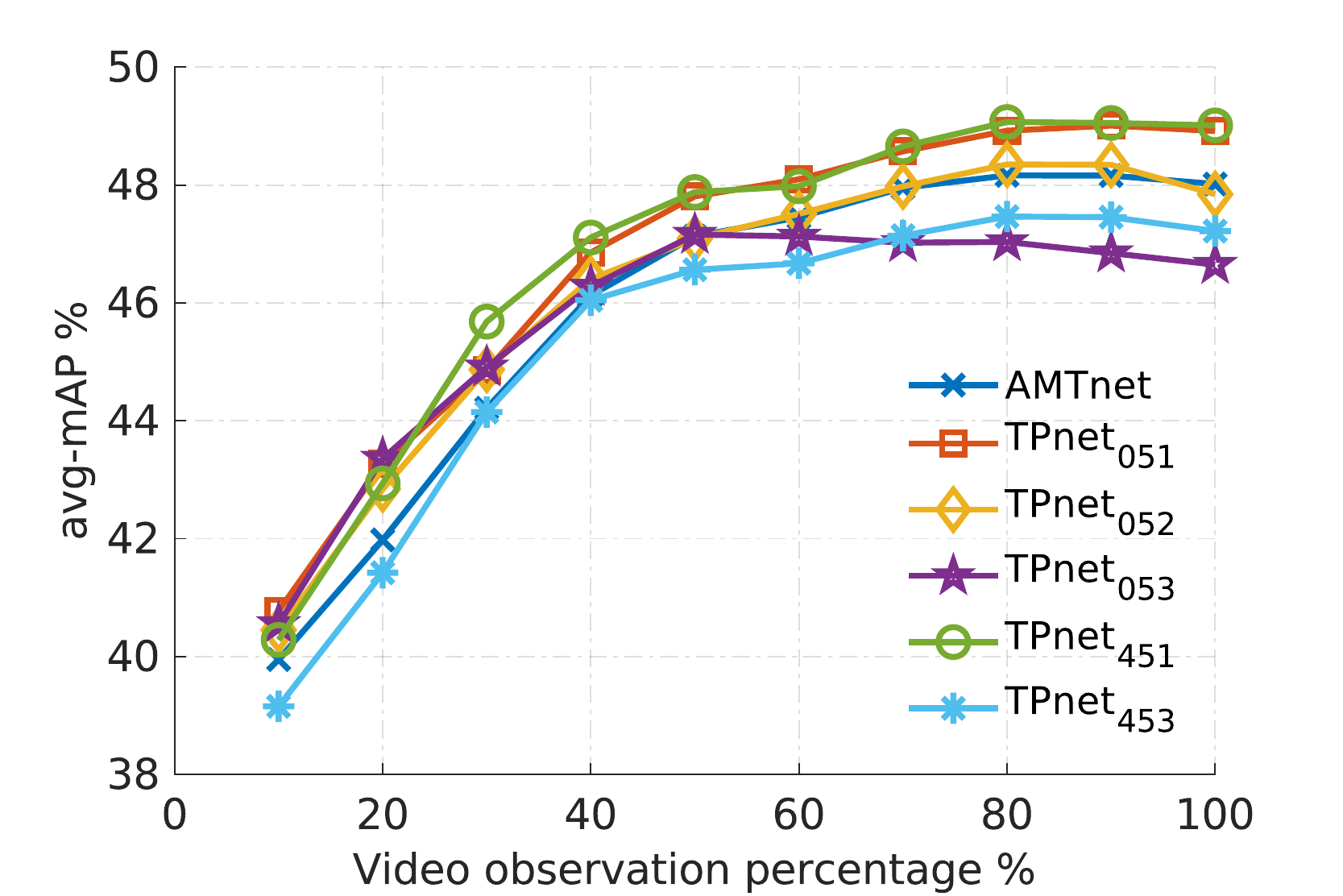}
  \caption{
    {\small
      \textit{Online action detection performance (avg-mAP) on J-HMDB-21 dataset.
      TPnet${}_{abc}$ represents our TPnet where $a = \Delta_p$, $b = \Delta_f$ and $c = n$.}
    }
 }
\label{fig:label}
\vspace{-10mm}
\end{figure*}

\begin{figure*}[h]
\vspace{-6mm}
  \centering
  \includegraphics[scale=0.60]{figures/supp/mAP@075_all.pdf}
  \caption{
    {\small
      \textit{Online action detection performance (mAP $\delta=0.75$) on J-HMDB-21 dataset.
      TPnet${}_{abc}$ represents our TPnet where $a = \Delta_p$, $b = \Delta_f$ and $c = n$}
    }
 }
\label{fig:label}
\vspace{-5mm}
\end{figure*}

\begin{figure*}[h]
\vspace{-3mm}
  \centering
  \includegraphics[scale=0.60]{figures/supp/mAP@05_online.pdf}
  \caption{
    {\small
      \textit{Online action detection performance (mAP $\delta=0.5$) on J-HMDB-21 dataset.
      TPnet${}_{abc}$ represents our TPnet where $a = \Delta_p$, $b = \Delta_f$ and $c = n$}
    }
 }
\label{fig:label}
\vspace{-10mm}
\end{figure*}

\begin{figure*}[h]
\vspace{-5mm}
  \centering
  \includegraphics[scale=0.60]{figures/supp/mAP@02_online.pdf}
  \caption{
    {\small
      \textit{Online action detection performance (mAP $\delta=0.2$) on J-HMDB-21 dataset.
      TPnet${}_{abc}$ represents our TPnet where $a = \Delta_p$, $b = \Delta_f$ and $c = n$}
    }
 }
\label{fig:label}
\vspace{-10mm}
\end{figure*}

\clearpage
\section{Action Tube Completion Performance (c-mAP).}
\begin{figure*}[h]
\vspace{-10mm}
  \centering
  \includegraphics[scale=0.60]{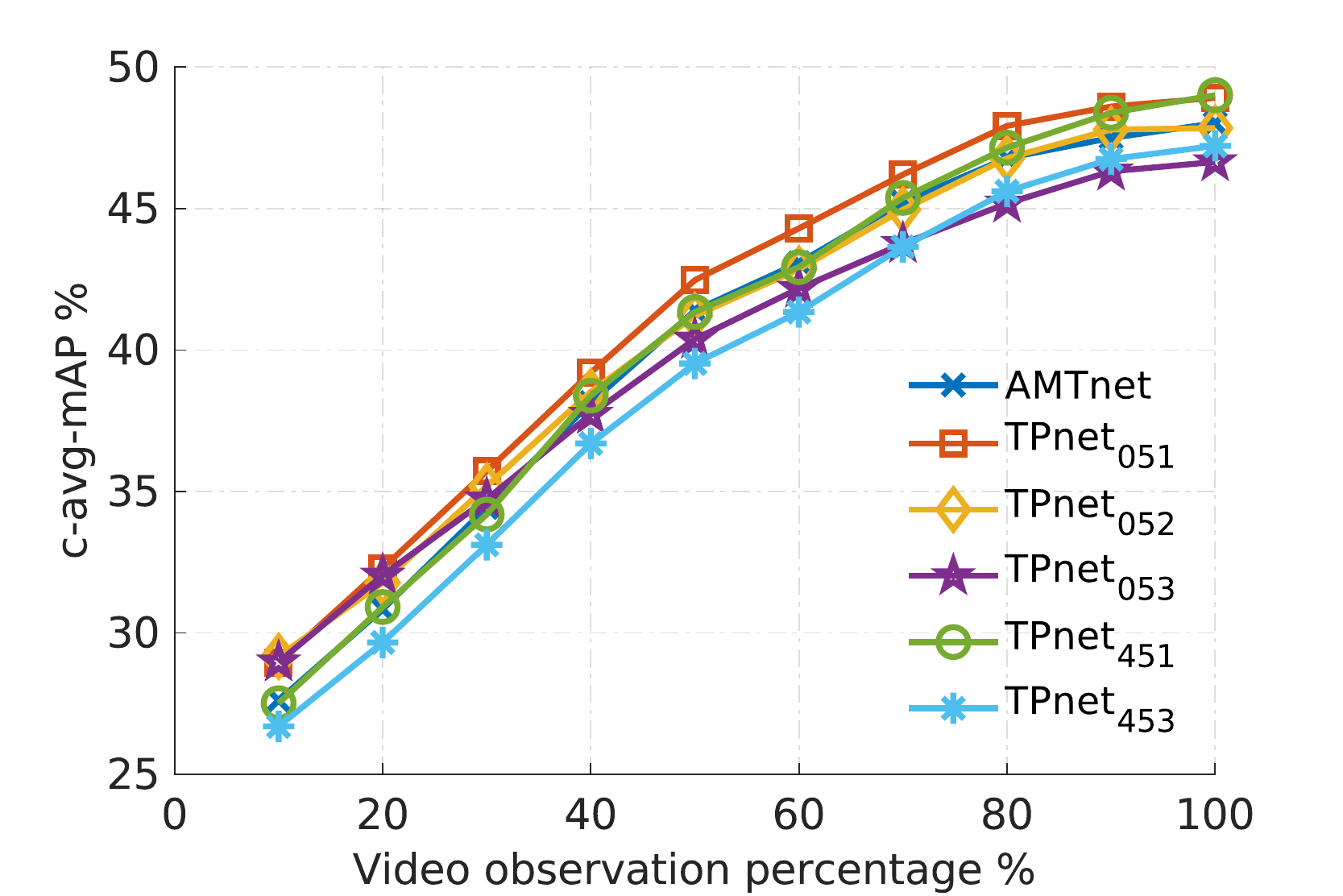}
  \caption{
    {\small
      \textit{Action tube completion performance (c-avg-mAP) on J-HMDB-21 dataset.
      TPnet${}_{abc}$ represents our TPnet where $a = \Delta_p$, $b = \Delta_f$ and $c = n$.}
    }
 }
\label{fig:label}
\vspace{-10mm}
\end{figure*}

\begin{figure*}[h]
\vspace{-6mm}
  \centering
  \includegraphics[scale=0.60]{figures/supp/mAP@075_all.pdf}
  \caption{
    {\small
      \textit{Action tube completion performance (c-mAP $\delta=0.75$) on J-HMDB-21 dataset.
      TPnet${}_{abc}$ represents our TPnet where $a = \Delta_p$, $b = \Delta_f$ and $c = n$}
    }
 }
\label{fig:label}
\vspace{-5mm}
\end{figure*}

\begin{figure*}[h]
\vspace{-3mm}
  \centering
  \includegraphics[scale=0.60]{figures/supp/mAP@05_all.pdf}
  \caption{
    {\small
      \textit{Action tube completion performance (c-mAP $\delta=0.5$) on J-HMDB-21 dataset.
      TPnet${}_{abc}$ represents our TPnet where $a = \Delta_p$, $b = \Delta_f$ and $c = n$}
    }
 }
\label{fig:label}
\vspace{-10mm}
\end{figure*}

\begin{figure*}[h]
\vspace{-5mm}
  \centering
  \includegraphics[scale=0.60]{figures/supp/mAP@02_all.pdf}
  \caption{
    {\small
      \textit{Action tube completion performance (c-mAP $\delta=0.2$) on J-HMDB-21 dataset.
      TPnet${}_{abc}$ represents our TPnet where $a = \Delta_p$, $b = \Delta_f$ and $c = n$}
    }
 }
\label{fig:label}
\vspace{-10mm}
\end{figure*}
\clearpage
\section{Future Action Tube Prediction Performance (p-mAP).}
\begin{figure*}[h]
\vspace{-10mm}
  \centering
  \includegraphics[scale=0.60]{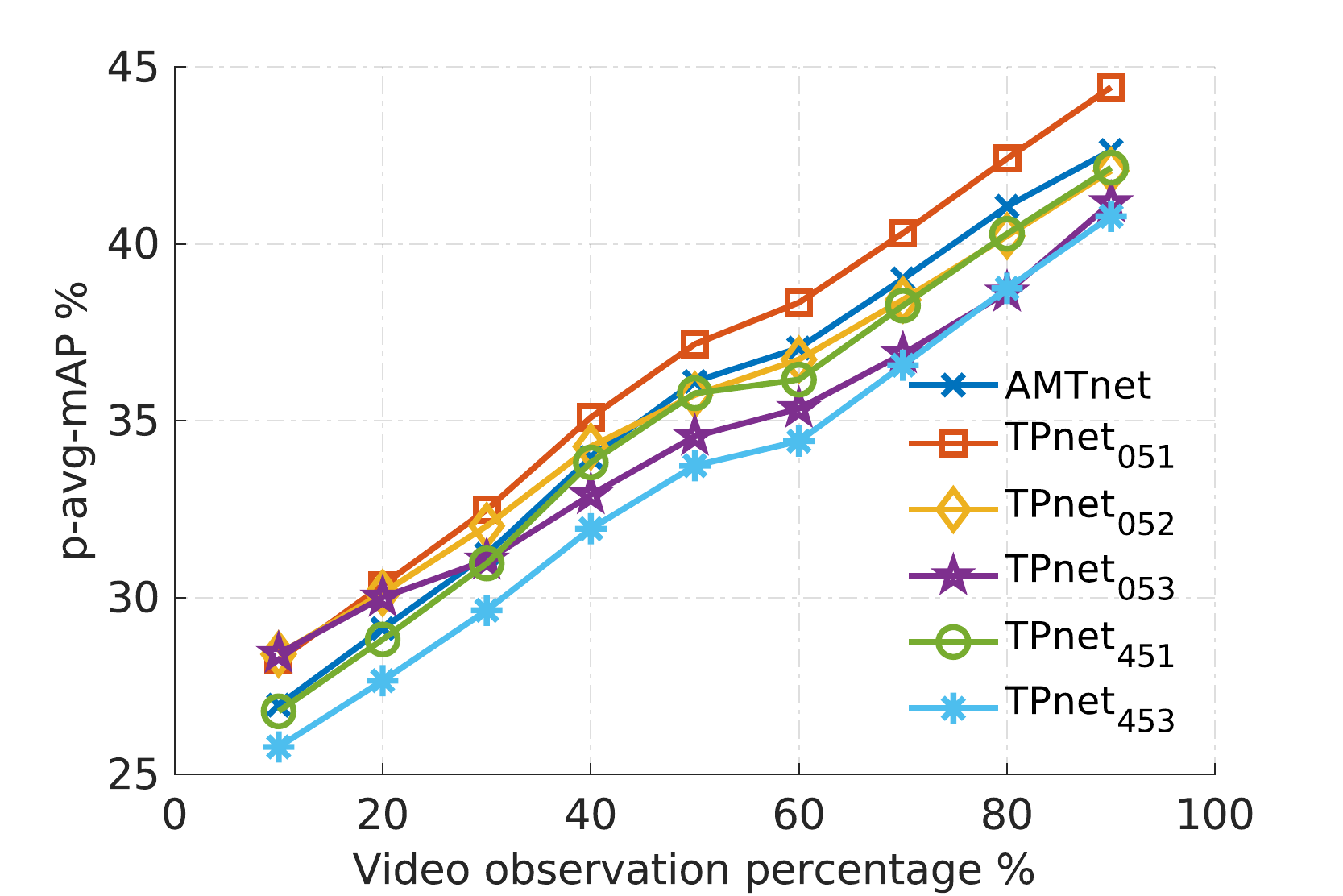}
  \caption{
    {\small
      \textit{Future action tube prediction performance (p-avg-mAP) on J-HMDB-21 dataset.
      TPnet${}_{abc}$ represents our TPnet where $a = \Delta_p$, $b = \Delta_f$ and $c = n$.}
    }
 }
\label{fig:label}
\vspace{-10mm}
\end{figure*}

\begin{figure*}[h]
\vspace{-6mm}
  \centering
  \includegraphics[scale=0.60]{figures/supp/mAP@075_predict.pdf}
  \caption{
    {\small
      \textit{Future action tube prediction performance (p-mAP $\delta=0.75$) on J-HMDB-21 dataset.
      TPnet${}_{abc}$ represents our TPnet where $a = \Delta_p$, $b = \Delta_f$ and $c = n$}
    }
 }
\label{fig:label}
\vspace{-5mm}
\end{figure*}

\begin{figure*}[h]
\vspace{-3mm}
  \centering
  \includegraphics[scale=0.60]{figures/supp/mAP@05_predict.pdf}
  \caption{
    {\small
      \textit{Future action tube prediction performance (p-mAP $\delta=0.5$) on J-HMDB-21 dataset.
      TPnet${}_{abc}$ represents our TPnet where $a = \Delta_p$, $b = \Delta_f$ and $c = n$}
    }
 }
\label{fig:label}
\vspace{-10mm}
\end{figure*}

\begin{figure*}[h]
\vspace{-5mm}
  \centering
  \includegraphics[scale=0.60]{figures/supp/mAP@02_predict.pdf}
  \caption{
    {\small
      \textit{Future action tube prediction performance (p-mAP $\delta=0.2$) on J-HMDB-21 dataset.
      TPnet${}_{abc}$ represents our TPnet where $a = \Delta_p$, $b = \Delta_f$ and $c = n$}
    }
 }
\label{fig:label}
\vspace{-10mm}
\end{figure*}